%% file: emnlp2017.tex
\documentclass[11pt,letterpaper]{article}
\usepackage{emnlp2017}
\usepackage{times}
\usepackage{latexsym}
\usepackage{times}
\usepackage{graphicx}
\usepackage{hhline}
\graphicspath{ {fig/} }
\usepackage{color}
\usepackage{float}
\usepackage{caption}
\usepackage{amssymb}
\usepackage{amsmath}
\usepackage{breqn}
\usepackage{verbatim}
\usepackage{microtype}
\usepackage{url}
\usepackage{array}
\usepackage{multirow}
\usepackage{makecell}
\usepackage{tabu}
\usepackage{pbox}
\usepackage{array}
\usepackage{verbatim}
\usepackage{tikz}
\usetikzlibrary{shapes,shapes.multipart}
\usepackage{xifthen}
\usetikzlibrary{decorations.pathreplacing}
\usetikzlibrary{matrix}
\usepackage[detect-all]{siunitx}
\usepackage{booktabs}
\usepackage{multirow}
\usepackage{etoolbox}
\robustify\bfseries

\usepackage{xcolor}
\usepackage{mdframed}

\emnlpfinalcopy 

\newcommand{\mnb}{Tensor Fusion Network}
\newcommand{\mns}{ TFN}

\DeclareMathOperator*{\argmax}{arg\,max}
\newcolumntype{L}{>{\centering\arraybackslash}m{6cm}}
\newcolumntype{M}{>{\centering\arraybackslash}m{1.6cm}}

\def\emnlppaperid{1400}

\title{\mnb \ for Multimodal Sentiment Analysis}

\author{%
 Amir Zadeh$^\dagger$, Minghai Chen$^\dagger$\\
 Language Technologies Institute \\
 Carnegie Mellon University \\
 {\tt \{abagherz,minghail\}@cs.cmu.edu} \\\And
 Soujanya Poria \\
 Temasek Laboratories, \\
 NTU, Singapore \\
 {\tt sporia@ntu.edu.sg} \\\AND
 Erik Cambria \\
 School of Computer Science and \\
Engineering, NTU, Singapore \\
 {\tt cambria@ntu.edu.sg} \\\And
 Louis-Philippe Morency \\
 Language Technologies Institute \\
 Carnegie Mellon University \\
 {\tt morency@cs.cmu.edu} \\}

\date{}

\begin{document}
\maketitle

\newcommand\blfootnote[1]{%
  \begingroup
  \renewcommand\thefootnote{}\footnote{#1}%
  \addtocounter{footnote}{-1}%
  \endgroup
}
\blfootnote{$^\dagger$ means equal contribution}

\begin{abstract}
Multimodal sentiment analysis is an increasingly popular research area, which extends the conventional language-based definition of sentiment analysis to a multimodal setup where other relevant modalities accompany language. In this paper, we pose the problem of multimodal sentiment analysis as modeling \textit{intra-modality} and \textit{inter-modality} dynamics. We introduce a novel model, termed \mnb, which learns both such dynamics end-to-end. 
The proposed approach is tailored for the volatile nature of spoken language in online videos as well as accompanying gestures and voice. In the experiments, our model outperforms state-of-the-art approaches for both multimodal and unimodal sentiment analysis.
\end{abstract}

\section{Introduction}

Multimodal sentiment analysis~\cite{morency2011towards,zadeh2016multimodal,poria2015deep} is an increasingly popular area of affective computing research~\cite{poria2017review} that focuses on generalizing text-based sentiment analysis to opinionated videos, where three communicative modalities are present: language (spoken words), visual (gestures), and acoustic (voice). 

This generalization is particularly vital to part of the NLP community dealing with opinion mining and sentiment analysis~\cite{campra} since there is a growing trend of sharing opinions in videos instead of text, specially in social media (Facebook, YouTube, etc.). 
The central challenge in multimodal sentiment analysis is to model the \textit{inter-modality} dynamics: the interactions between language, visual and acoustic behaviors that change the perception of the expressed sentiment. 

\input{fig/motivation.tex}

Figure~\ref{fig:motivation} illustrates these complex inter-modality dynamics. The utterance ``This movie is sick'' can be ambiguous (either positive or negative) by itself, but if the speaker is also smiling at the same time, then it will be perceived as positive. On the other hand, the same utterance with a frown would be perceived negatively. A person speaking loudly ``This movie is sick'' would still be ambiguous. These examples are illustrating \textbf{bimodal} interactions. Examples of \textbf{trimodal} interactions are shown in Figure~\ref{fig:motivation} when loud voice increases the sentiment to strongly positive. The complexity of inter-modality dynamics is shown in the second trimodal example where the utterance ``This movie is fair'' is still weakly positive, given the strong influence of the word ``fair''. 

A second challenge in multimodal sentiment analysis is efficiently exploring \textit{intra-modality} dynamics of a specific modality (\textbf{unimodal} interaction). Intra-modality dynamics are particularly challenging for the language analysis since multimodal sentiment analysis is performed on spoken language. A spoken opinion such as ``I think it was alright \dots Hmmm \dots let me think \dots yeah \dots no \dots ok yeah'' almost never happens in written text. This volatile nature of spoken opinions, where proper language structure is often ignored, complicates sentiment analysis. Visual and acoustic modalities also contain their own intra-modality dynamics which are expressed through both space and time.

Previous works in multimodal sentiment analysis does not account for both intra-modality and inter-modality dynamics directly, instead they either perform early fusion (a.k.a., feature-level fusion) or late fusion (a.k.a., decision-level fusion). Early fusion consists in simply concatenating multimodal features mostly at input level~\cite{morency2011towards,perez2013utterance,poria2016convolutional}. This fusion approach does not allow the intra-modality dynamics to be efficiently modeled. This is due to the fact that inter-modality dynamics can be more complex at input level and can dominate the learning process or result in overfitting. Late fusion, instead, consists in training unimodal classifiers independently and performing decision voting~\cite{wang2016select,zadeh2016mosi}. This prevents the model from learning inter-modality dynamics in an efficient way by assuming that simple weighted averaging is a proper fusion approach.

In this paper, we introduce a new model, termed \mnb \ (TFN), which learns both the intra-modality and inter-modality dynamics end-to-end. Inter-modality dynamics are modeled with a new multimodal fusion approach, named Tensor Fusion, which explicitly aggregates unimodal, bimodal and trimodal interactions. Intra-modality dynamics are modeled through three Modality Embedding Subnetworks, for language, visual and acoustic modalities, respectively. 

In our extensive set of experiments, we show (a) that TFN outperforms previous state-of-the-art approaches for multimodal sentiment analysis, (b) the characteristics and capabilities of our Tensor Fusion approach for multimodal sentiment analysis, and (c) that each of our three Modality Embedding Subnetworks (language, visual and acoustic) are also outperforming unimodal state-of-the-art unimodal sentiment analysis approaches.

\section{Related Work}

\textit{Sentiment Analysis} is a well-studied research area in NLP~\cite{pang2008opinion}. Various approaches have been proposed to model sentiment from language, including methods that focus on opinionated words~\cite{hu2004mining, taboada2011lexicon, poria2014sentic,camnt4}, $n$-grams and language models~\cite{yang2012extracting}, sentiment compositionality and dependency-based analysis~\cite{socher2013recursive, poria2014dependency, agarwal2015concept, tai2015improved}, and distributional representations for sentiment~\cite{iyyer2015deep}.

\textit{Multimodal Sentiment Analysis} is an emerging research area that integrates verbal and nonverbal behaviors into the detection of user sentiment. There exist several multimodal datasets that include sentiment annotations, including the newly-introduced CMU-MOSI dataset~\cite{zadeh2016multimodal}, as well as other datasets including ICT-MMMO~\cite{wollmer2013youtube}, YouTube~\cite{morency2011towards}, and MOUD~\cite{perez2013utterance}, however CMU-MOSI is the only English dataset with utterance-level sentiment labels. The newest multimodal sentiment analysis approaches have used deep neural networks, including convolutional neural networks (CNNs) with multiple-kernel learning~\cite{poria2015deep}, SAL-CNN~\cite{wang2016select} which learns generalizable features across speakers, and support vector machines (SVMs) with a multimodal dictionary~\cite{zadeh2015micro}.

\textit{Audio-Visual Emotion Recognition} is closely tied to multimodal sentiment analysis~\cite{poria2017review}. Both audio and visual features have been shown to be useful in the recognition of emotions~\cite{Ghosh+2016}. Using facial expressions and audio cues jointly has been the focus of many recent studies~\cite{glodek2011multiple, valstar2016avec, nojavanasghari2016emoreact}. 

\textit{Multimodal Machine Learning} has been a growing trend in machine learning research that is closely tied to the studies in this paper. Creative and novel applications of using multiple modalities have been among successful recent research directions in machine learning \cite{you2016image, donahue2015long,antol2015vqa,specia2016shared,traffickingacl17}. 

\section{CMU-MOSI Dataset}\label{sec:dataset}
\input{fig/mosi_graph.tex}
Multimodal Opinion Sentiment Intensity (CMU-MOSI) dataset is an annotated dataset of video opinions from YouTube movie reviews~\cite{zadeh2016mosi}. Annotation of sentiment has closely followed the annotation scheme of the Stanford Sentiment Treebank~\cite{socher2013recursive}, where sentiment is annotated on a seven-step Likert scale from very negative to very positive. 
However, whereas the Stanford Sentiment Treebank is segmented by sentence, the CMU-MOSI dataset is segmented by opinion utterances to accommodate spoken language where sentence boundaries are not as clear as text. There are 2199 opinion utterances for 93 distinct speakers in CMU-MOSI. There are an average 23.2 opinion segments in each video. Each video has an average length of 4.2 seconds. There are a total of 26,295 words in the opinion utterances. These utterance are annotated by five Mechanical Turk annotators for sentiment. The final agreement between the annotators is high in terms of Krippendorf's alpha $\alpha = 0.77$. Figure~\ref{fig:mosi_graph} shows the distribution of sentiment across different opinions and different opinion sizes. CMU-MOSI dataset facilitates three prediction tasks, each of which we address in our experiments: 1) \textit{Binary Sentiment Classification} 2) \textit{Five-Class Sentiment Classification} (similar to Stanford Sentiment Treebank fine-grained classification with seven scale being mapped to five) and 3) \textit{Sentiment Regression} in range $[-3,3]$. For sentiment regression, we report Mean-Absolute Error (lower is better) and correlation (higher is better) between the model predictions and regression ground truth.

\section{\mnb}
Our proposed TFN consists of three major components: 1) \textit{Modality Embedding Subnetworks} take as input unimodal features, and output a rich modality embedding. 2) \textit{Tensor Fusion Layer} explicitly models the unimodal, bimodal and trimodal interactions using a 3-fold Cartesian product from modality embeddings. 3) \textit{Sentiment Inference Subnetwork} is a network conditioned on the output of the Tensor Fusion Layer and performs sentiment inference. Depending on the task from Section~\ref{sec:dataset} the network output changes to accommodate binary classification, 5-class classification or regression. 
Input to the TFN is an opinion utterance which includes three modalities of language, visual and acoustic. The following three subsections describe the TFN subnetworks and their inputs in detail. 

\subsection{Modality Embedding Subnetworks}\label{subsec:unimodal}

\textbf{Spoken Language Embedding Subnetwork:} 
Spoken text is different than written text (reviews, tweets) in compositionality and grammar. We revisit the spoken opinion: ``I think it was alright \dots Hmmm \dots let me think \dots yeah \dots no \dots ok yeah''. This form of opinion rarely happens in written language but variants of it are very common in spoken language. The first part conveys the actual message and the rest is speaker thinking out loud eventually agreeing with the first part. The key factor in dealing with this volatile nature of spoken language is to build models that are capable of operating in presence of unreliable and idiosyncratic speech traits by focusing on important parts of speech. 

Our proposed approach to deal with challenges of spoken language is to learn a rich representation of spoken words at each word interval and use it as input to a fully connected deep network (Figure~\ref{fig:lang}). This rich representation for $i$th word contains information from beginning of utterance through time, as well as $i$th word. This way as the model is discovering the meaning of the utterance through time, if it encounters unusable information in word $i+1$ and arbitrary number of words after, the representation up until $i$ is not diluted or lost. Also, if the model encounters usable information again, it can recover by embedding those in the long short-term memory (LSTM). The time-dependent encodings are usable by the rest of the pipeline by simply focusing on relevant parts using the non-linear affine transformation of time-dependent embeddings which can act as a dimension reducing attention mechanism. To formally define our proposed Spoken Language Embedding Subnetwork ($\mathcal{U}_l$), let $\mathbf{l}=\{l_1,l_2,l_3,\dots,l_{T_l} ; l_t \in \mathbb{R}^{300} \}$, where $T_l$ is the number of words in an utterance, be the set of spoken words represented as a sequence of 300-dimensional GloVe word vectors~\cite{pennington2014glove}. 

A LSTM network~\cite{hochreiter1997long} with a forget gate~\cite{gers2000learning} is used to learn time-dependent language representations $\mathbf{h_l}=\{h_1,h_2,h_3,\dots,h_{T_l} ; h_t \in \mathbb{R}^{128} \}$ for words according to the following LSTM formulation.
\begin{align*}
 \begin{pmatrix} i \\ f \\ o \\ m \end{pmatrix} &= 
 \begin{pmatrix} sigmoid \\ sigmoid \\ sigmoid \\ tanh \end{pmatrix} W_{l_d} \begin{pmatrix} X_t W_{l_e} \\ h_{t-1} \end{pmatrix} \\
 c_t&=f \odot c_{t-1} + i \odot m \\
 h_t&=o \otimes tanh (c_t) \\
 \mathbf{h_l}&=[h_1;h_2;h_3; \dots; h_{T_l}]
\end{align*}
$\mathbf{h_l}$ is a matrix of language representations formed from concatenation of $h_1, h_2, h_3, \dots h_{T_l}$. $\mathbf{h_l}$ is then used as input to a fully-connected network that generates language embedding $\mathbf{z}^{l}$:
\begin{equation*}
	\mathbf{z}^{l}=\mathcal{U}_l(\mathbf{l};\ W_l) \in \mathbb{R}^{128}
\end{equation*}
where $W_l$ is the set of all weights in the $\mathcal{U}_l$ network (including $W_{l_d}$, $W_{l_e}$,$W_{l_{fc}}$, and $b_{l_{fc}}$), $\sigma$ is the sigmoid function. 
\input{fig/language.tex}

\textbf{Visual Embedding Subnetwork:} Since opinion videos consist mostly of speakers talking to the audience through close-up camera, face is the most important source of visual information. The speaker's face is detected for each frame (sampled at 30Hz) and indicators of the seven basic emotions (anger, contempt, disgust, fear, joy, sadness, and surprise) and two advanced emotions (frustration and confusion)~\cite{ekman1992argument} are extracted using FACET facial expression analysis framework\footnote{\url{http://goo.gl/1rh1JN}}. A set of 20 Facial Action Units~\cite{ekman1980facial}, indicating detailed muscle movements on the face, are also extracted using FACET. Estimates of head position, head rotation, and 68 facial landmark locations also extracted per frame using OpenFace~\cite{baltruvsaitis2016openface,ceclm17}.

Let the visual features $\hat{\mathbf{v}}_j=[v_j^1,v_j^2,v_j^3,\dots, v_j^{p}]$ for frame $j$ of utterance video contain the set of $p$ visual features, with $T_v$ the number of total video frames in utterance. We perform mean pooling over the frames to obtain the expected visual features $\mathbf{v}=[\mathbb{E}[v^1], \mathbb{E}[v^2], \mathbb{E}[v^3], \dots, \mathbb{E}[v^l]]$. $\mathbf{v}$ is then used as input to the Visual Embedding Subnetwork $\mathcal{U}_v$. Since information extracted using FACET from videos is rich, using a deep neural network would be sufficient to produce meaningful embeddings of visual modality. We use a deep neural network with three hidden layers of 32 ReLU units and weights $W_v$. Empirically we observed that making the model deeper or increasing the number of neurons in each layer does not lead to better visual performance. The subnetwork output provides the visual embedding $\mathbf{z}^{v}$:
\begin{equation*}
\mathbf{z}^{v}=\mathcal{U}_v(\mathbf{v};\ W_v)\in \mathbb{R}^{32}
\end{equation*}

\input{fig/fusion_fig.tex}

\textbf{Acoustic Embedding Subnetwork:}
For each opinion utterance audio, a set of acoustic features are extracted using COVAREP acoustic analysis framework~\cite{degottex2014covarep}, including 12 MFCCs, pitch tracking and Voiced/UnVoiced segmenting features (using the additive noise robust \textit{Summation of Residual Harmonics} (SRH) method~\cite{drugman2011joint}), glottal source parameters (estimated by glottal inverse filtering based on GCI synchronous IAIF~\cite{drugman2012detection, alku1992glottal, alku2002normalized, alku1997parabolic, titze1992vocal, childers1991vocal}), peak slope parameters~\cite{degottex2014covarep}, maxima dispersion quotients (MDQ)~\cite{kane2013wavelet}, and estimations of the $R_d$ shape parameter of the Liljencrants-Fant (LF) glottal model~\cite{fujisaki1986proposal}. These extracted features capture different characteristics of human voice and have been shown to be related to emotions~\cite{ghosh2016representation}. 

For each opinion segment with $T_a$ audio frames (sampled at 100Hz; i.e., 10ms), we extract the set of $q$ acoustic features $\hat{\mathbf{a}_j}=[a_j^1,a_j^2,a_j^3,\dots, a_j^{q}]$ for audio frame $j$ in utterance. We perform mean pooling per utterance on these extracted acoustic features to obtain the expected acoustic features $\mathbf{a}=[\mathbb{E}[a_1], \mathbb{E}[a_2], \mathbb{E}[a_3], \dots, \mathbb{E}[q]]$. Here, $\mathbf{a}$ is the input to the Audio Embedding Subnetwork $\mathcal{U}_a$. Since COVAREP also extracts rich features from audio, using a deep neural network is sufficient to model the acoustic modality. Similar to $\mathcal{U}_v$, $\mathcal{U}_a$ is a network with 3 layers of 32 ReLU units with weights $W_a$. 

Here, we also empirically observed that making the model deeper or increasing the number of neurons in each layer does not lead to better performance. The subnetwork produces the audio embedding $\mathbf{z}^{a}$: 
\begin{equation*}
\mathbf{z}^{a}=\mathcal{U}_a(\mathbf{a}; W_a) \in \mathbb{R}^{32}
\end{equation*}

\subsection{Tensor Fusion Layer}
While previous works in multimodal research has used feature concatenation as an approach for multimodal fusion, we aim to build a fusion layer in TFN that disentangles unimodal, bimodal and trimodal dynamics by modeling each of them explicitly. We call this layer Tensor Fusion, which is defined as the following vector field using three-fold Cartesian product:
\begin{align*}
	\bigg\{(z^l, z^v, z^a)\ | \ z^l \in \begin{bmatrix} \mathbf{z}^{l} \\ 1 \end{bmatrix}, 
		z^v \in \begin{bmatrix} \mathbf{z}^{v} \\ 1 \end{bmatrix},
		z^a \in \begin{bmatrix} \mathbf{z}^{a} \\ 1 \end{bmatrix} \bigg\}
\end{align*}
The extra constant dimension with value 1 generates the unimodal and bimodal dynamics. Each neural coordinate $(z_l, z_v, z_a)$ can be seen as a 3-D point in the 3-fold Cartesian space defined by the language, visual, and acoustic embeddings dimensions $[\mathbf{z}^l 1]^T$, $[\mathbf{z}^v 1]^T$, and $[\mathbf{z}^a 1]^T$.

This definition is mathematically equivalent to a differentiable outer product between $\mathbf{z}^{l}$, the visual representation $\mathbf{z}^{v}$, and the acoustic representation $\mathbf{z}^{a}$.
\begin{equation*}
	\mathbf{z}^m = \begin{bmatrix} \mathbf{z}^{l} \\ 1 \end{bmatrix}
		\otimes \begin{bmatrix} \mathbf{z}^{v} \\ 1 \end{bmatrix}
  \otimes \begin{bmatrix} \mathbf{z}^{a} \\ 1 \end{bmatrix} 
\end{equation*} 
Here $\otimes$ indicates the outer product between vectors and $\mathbf{z}^m \in \mathbb{R}^{129 \times 33 \times 33}$ is the 3D cube of all possible combination of unimodal embeddings with seven semantically distinct subregions in Figure~\ref{fig:fusion}. The first three subregions $\mathbf{z}^{l}$, $\mathbf{z}^{v}$, and $\mathbf{z}^{a}$ are unimodal embeddings from Modality Embedding Subnetworks forming unimodal interactions in Tensor Fusion. Three subregions $\mathbf{z}^{l} \otimes \mathbf{z}^{v}$, $\mathbf{z}^{l} \otimes \mathbf{z}^{a}$, and 
$\mathbf{z}^{v} \otimes \mathbf{z}^{a}$ capture bimodal interactions in Tensor Fusion. Finally, $\mathbf{z}^{l} \otimes \mathbf{z}^{v} \otimes \mathbf{z}^{a}$ captures trimodal interactions. 

Early fusion commonly used in multimodal research dealing with language, vision and audio, can be seen as a special case of Tensor Fusion with only unimodal interactions. Since Tensor Fusion is mathematically formed by an outer product, it has no learnable parameters and we empirically observed that although the output tensor is high dimensional, chances of overfitting are low. 

We argue that this is due to the fact that the output neurons of Tensor Fusion are easy to interpret and semantically very meaningful (i.e., the manifold that they lie on is not complex but just high dimensional). Thus, it is easy for the subsequent layers of the network to decode the meaningful information. 
\subsection{Sentiment Inference Subnetwork}
After Tensor Fusion layer, each opinion utterance can be represented as a multimodal tensor $\mathbf{z}^m$. We use a fully connected deep neural network called Sentiment Inference Subnetwork $\mathcal{U}_s$ with weights $W_s$ conditioned on $\mathbf{z}^m$. The architecture of the network consists of two layers of 128 ReLU activation units connected to decision layer. The likelihood function of the Sentiment Inference Subnetwork is defined as follows, where $\boldsymbol{\phi}$ is the sentiment prediction:
\begin{equation*}
\argmax_{\boldsymbol{\phi}}\ \textrm{p}(\boldsymbol{\phi} \ |\ \mathbf{z}^m; W_s) = \argmax_{\boldsymbol{\phi}}\ \mathcal{U}_s(\mathbf{z}^m ; W_s)
\label{eq:SIN}
\end{equation*}
In our experiments, we use three variations of the $\mathcal{U}_s$ network. The first network is trained for binary sentiment classification, with a single sigmoid output neuron using binary cross-entropy loss. The second network is designed for five-class sentiment classification, and uses a softmax probability function using categorical cross-entropy loss. The third network uses a single sigmoid output, using mean-squarred error loss to perform sentiment regression.

\section{Experiments}\label{sec:experiments}
In this paper, we devise three sets of experiments each addressing a different research question:

\textbf{Experiment 1}: We compare our TFN with previous state-of-the-art approaches in multimodal sentiment analysis. 

\textbf{Experiment 2}: We study the importance of the TFN subtensors and the impact of each individual modality (see Figure~\ref{fig:fusion}). We also compare with the commonly-used early fusion approach. 

\textbf{Experiment 3}: We compare the performance of our three modality-specific networks (language, visual and acoustic) with state-of-the-art unimodal approaches. 

Section~\ref{sec:methodology} describes our experimental methodology which is kept constant across all experiments. Section~\ref{sec:discussion} will discuss our results in more details with a qualitative analysis.

\subsection{E1: Multimodal Sentiment Analysis}
\label{sec:ex1}
In this section, we compare the performance of TFN model with previously proposed multimodal sentiment analysis models. We compare to the following baselines: 

\textbf{C-MKL}~\cite{poria2015deep} Convolutional MKL-based model is a multimodal sentiment classification model which uses a CNN to extract textual features and uses multiple kernel learning for sentiment analysis. It is current SOTA (state of the art) on CMU-MOSI. 

\textbf{SAL-CNN}~\cite{wang2016select} Select-Additive Learning is a multimodal sentiment analysis model that attempts to prevent identity-dependent information from being learned in a deep neural network. We retrain the model for 5-fold cross-validation using the code provided by the authors on github. 

\textbf{SVM-MD} ~\cite{zadeh2016multimodal} is a SVM model trained on multimodal features using early fusion. The model used in~\cite{morency2011towards} and~\cite{perez2013utterance} also similarly use SVM on multimodal concatenated features. We also present the results of Random Forest \textbf{RF-MD} to compare to another non-neural approach. 

The results first experiment are reported in Table~\ref{table:mmres}. TFN outperforms previously proposed neural and non-neural approaches. This difference is specifically visible in the case of 5-class classification.

\subsection{E2: Tensor Fusion Evaluation} 

Table~\ref{fig:fusion} shows the results of our ablation study. The first three rows are showing the performance of each modality, when no intermodality dynamics are modeled. From this first experiment, we observe that the language modality is the most predictive. 

\input{tables/ex1_table.tex}

As a second set of ablation experiments, we test our TFN approach when only the bimodal subtensors are used (TFN$_{bimodal}$) or when only the trimodal subtensor is used (TFN$_{bimodal}$). We observe that bimodal subtensors are more informative when used without other subtensors. The most interesting comparison is between our full TFN model and a variant (TFN$_{notrimodal}$) where the trimodal subtensor is removed (but all the unimodal and bimodal subtensors are present). We observe a big improvement for the full TFN model, confirming the importance of the trimodal dynamics and the need for all components of the full tensor. 

\input{tables/ex2_table2.tex}

We also perform a comparison with the early fusion approach (TFN$_{early}$) by simply concatenating all three modality embeddings $<z^l, z^a, z^v>$ and passing it directly as input to $\mathcal{U}_s$. This approach was depicted on the left side of Figure~\ref{fig:fusion}. When looking at Table~\ref{fig:fusion} results, we see that our TFN approach outperforms the early fusion approach\footnote{We also performed other comparisons with variants of the early fusion model \mns$_{early}$ where we increased the number of parameters and neurons to replicate the numbers from our TFN model. In all cases, the performances were similar to \mns$_{early}$ (and lower than our TFN model). Because of space constraints, we could not include them in this paper.}.

\input{tables/ex3_table_l.tex}


\subsection{E3: Modality Embedding Subnetworks Evaluation}
In this experiment, we compare the performance of our Modality Embedding Networks with state-of-the-art approaches for language-based, visual-based and acoustic-based sentiment analysis.

\subsubsection{Language Sentiment Analysis}

We selected the following state-of-the-art approaches to include variety in their techniques, based on dependency parsing (RNTN), distributional representation of text (DAN), and convolutional approaches (DynamicCNN).  
When possible, we retrain them on the CMU-MOSI dataset (performances of the original pre-trained models are shown in parenthesis in Table~\ref{table:lres}) and compare them to our language only \mns$_{language}$. 

\textbf{RNTN}~\cite{socher2013recursive}The Recursive Neural Tensor Network is among the most well-known sentiment analysis methods proposed for both binary and multi-class sentiment analysis that uses dependency structure. 

\textbf{DAN} ~\cite{iyyer2015deep} The Deep Average Network approach is a simple but efficient sentiment analysis model that uses information only from distributional representation of the words and not from the compositionality of the sentences.

\textbf{DynamicCNN} ~\cite{kalchbrenner2014convolutional} DynamicCNN is among the state-of-the-art models in text-based sentiment analysis which uses a convolutional architecture adopted for the semantic modeling of sentences.

\textbf{CMK-L}, \textbf{SAL-CNN-L} and \textbf{SVM-MD-L} are multimodal models from section using only language modality~\ref{sec:ex1}.

Results in Table~\ref{table:lres} show that our model using only language modality outperforms state-of-the-art approaches for the CMU-MOSI dataset. While previous models are well-studied and suitable models for sentiment analysis in written language, they underperform in modeling the sentiment in spoken language. We suspect that this underperformance is due to: RNTN and similar approaches rely heavily on dependency structure, which may not be present in spoken language; DAN and similar sentence embeddings approaches can easily be diluted by words that may not relate directly to sentiment or meaning; D-CNN and similar convolutional approaches rely on spatial proximity of related words, which may not always be present in spoken language. 

\input{tables/ex3_table_2.tex}

\subsubsection{Visual Sentiment Analysis}
We compare the performance of our models using visual information (TFN$_{visual}$) with the following well-known approaches in visual sentiment analysis and emotion recognition (retrained for sentiment analysis):

\textbf{3DCNN}~\cite{byeon2014facial} a network using 3D CNN is trained using the face of the speaker. Face of the speaker is extracted in every 6 frames and resized to $64 \times 64$ and used as the input to the proposed network. 

\textbf{CNN-LSTM}~\cite{ebrahimi2015recurrent} is a recurrent model that at each timestamp performs convolutions over facial region and uses output to an LSTM. Face processing is similar to 3DCNN. 

\textbf{LSTM-FA} similar to both baselines above, information extracted by FACET is used every 6 frames as input to an LSTM with a memory dimension of 100 neurons. 

\textbf{SAL-CNN-V}, \textbf{SVM-MD-V}, \textbf{CMKL-V}, \textbf{RF-V} use only visual modality in multimodal baselines from Section~\ref{sec:ex1}.

The results in Table~\ref{table:vres} show that $\mathcal{U}_v$ is able to outperform state-of-the-art approaches on visual sentiment analysis.

\input{tables/ex3_table_3.tex}

\subsubsection{Acoustic Sentiment Analysis}
We compare the performance of our models using visual information (TFN$_{acoustic}$) with the following well-known approaches in audio sentiment analysis and emotion recognition (retrained for sentiment analysis):

\textbf{HL-RNN} ~\cite{lee2015high} uses an LSTM on high-level audio features. We use the same features extracted for $\mathcal{U}_a$ averaged over time slices of every 200 intervals. 

\textbf{Adieu-Net}~\cite{trigeorgis2016adieu} is an end-to-end approach for emotion recognition in audio using directly PCM features. 

\textbf{SER-LSTM}~\cite{lim2016speech} is a model that uses recurrent neural networks on top of convolution operations on spectrogram of audio. 

\textbf{SAL-CNN-A}, \textbf{SVM-MD-A}, \textbf{CMKL-A}, \textbf{RF-A} use only acoustic modality in multimodal baselines from Section~\ref{sec:ex1}.

\input{tables/ComparisonTable.tex}

\subsection{Methodology}\label{sec:methodology}
All the models in this paper are tested using five-fold cross-validation proposed by CMU-MOSI~\cite{zadeh2016mosi}. All of our experiments are performed independent of speaker identity, as no speaker is shared between train and test sets for generalizability of the model to unseen speakers in real-world. The best hyperparameters are chosen using grid search based on model performance on a validation set (using last 4 videos in train fold). The TFN model is trained using the Adam optimizer~\cite{kingma2014adam} with the learning rate $5\mathrm{e}−4$. $\mathcal{U}_v$ and $\mathcal{U}_a$, $\mathcal{U}_s$ subnetworks are regularized using dropout on all hidden layers with $p=0.15$ and L2 norm coefficient $0.01$. The train, test and validation folds are exactly the same for all baselines.

\section{Qualitative Analysis}\label{sec:discussion}
We analyze the impact of our proposed TFN multimodal fusion approach by comparing it with the early fusion approach \mns$_{early}$ and the three unimodal models. Table~\ref{table:comparison} shows examples taken from the CMU-MOSI dataset. Each example is described with the spoken words as well as the acoustic and visual behaviors. The sentiment predictions and the ground truth labels range between strongly negative (-3) and strongly positive (+3). 

As a first general observation, we observe that the early fusion model \mns$_{early}$ shows a strong preference for the language modality and seems to be neglecting the intermodality dynamics. 
We can see this trend by comparing it with the language unimodal model \mns$_{language}$. In comparison, our TFN approach seems to capture more complex interaction through bimodal and trimodal dynamics and thus performs better. 
Specifically, in the first example, the utterance is weakly negative where the speaker is referring to lack of funny jokes in the movie. This example contains a bimodal interaction where the visual modality shows a negative expression (frowning) which is correctly captured by our TFN approach. 

In the second example, the spoken words are ambiguous since the model has no clue what a B is except a token, but the acoustic and visual modalities are bringing complementary evidences. Our TFN approach correctly identify this trimodal interaction and predicts a positive sentiment. The third example is interesting since it shows an interaction where language predicts a positive sentiment but the strong negative visual behaviors bring the final prediction of our TFN approach almost to a neutral sentiment. The fourth example shows how the acoustic modality is also influencing our TFN predictions.

\section{Conclusion}
We introduced a new end-to-end fusion method for sentiment analysis which explicitly represents unimodal, bimodal, and trimodal interactions between behaviors. Our experiments on the publicly-available CMU-MOSI dataset produced state-of-the-art performance when compared against both multimodal approaches. Furthermore, our approach brings state-of-the-art results for language-only, visual-only and acoustic-only multimodal sentiment analysis on CMU-MOSI. 

\section*{Acknowledgments}
This project was partially supported by Oculus research grant. We would like to thank the reviewers for their valuable feedback. 

\bibliography{emnlp2017,sentic}
\bibliographystyle{emnlp_natbib}

\end{document}

%% file: fig/motivation.tex
\begin{figure}[t]
	\centering
{\includegraphics[width=0.95\linewidth]{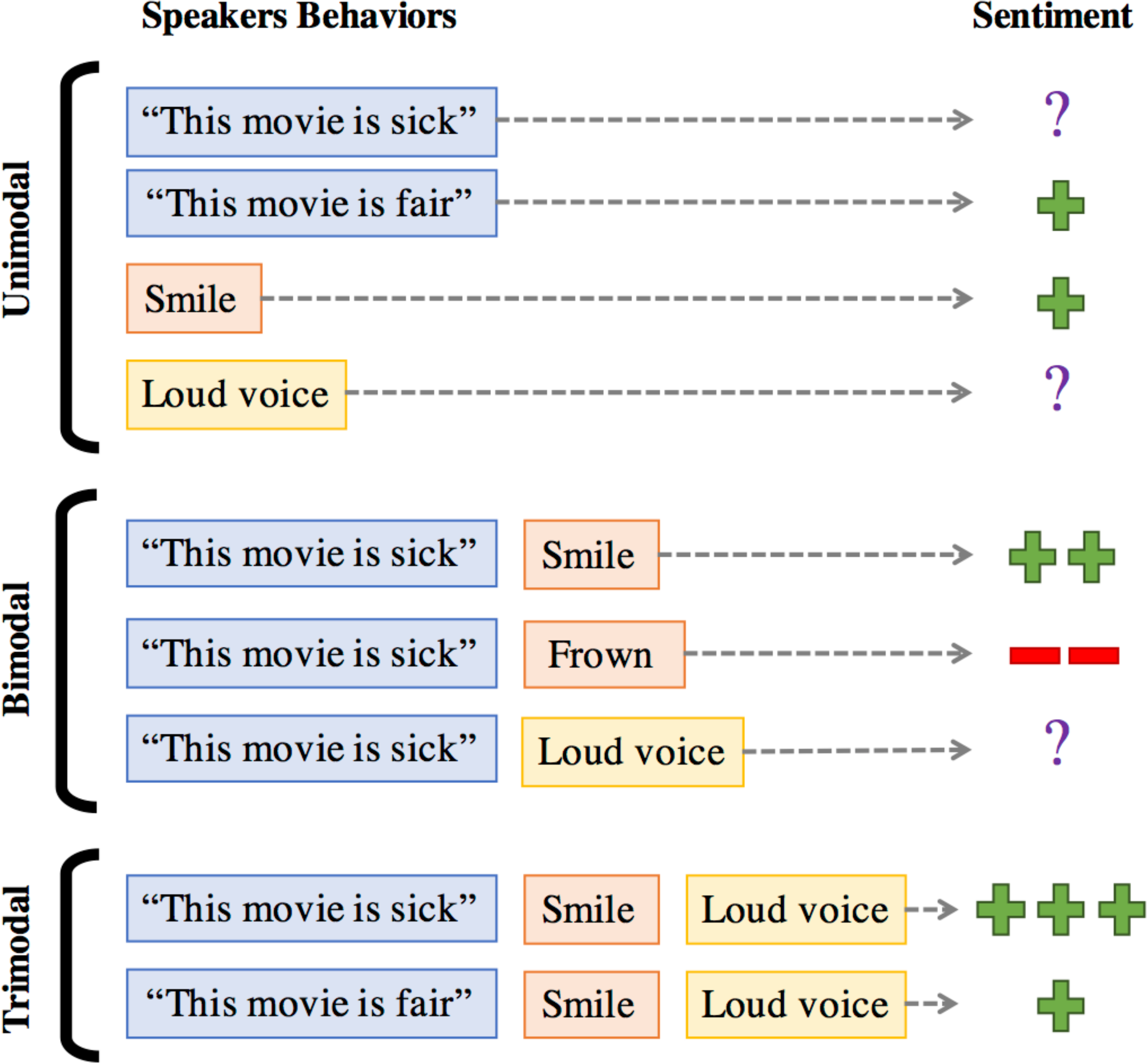}} 
    \caption{Unimodal, bimodal and trimodal interaction in multimodal sentiment analysis.}
    \label{fig:motivation}
\end{figure}

%% file: fig/mosi_graph.tex
\begin{figure*}[t]
	\centering
{\includegraphics[]{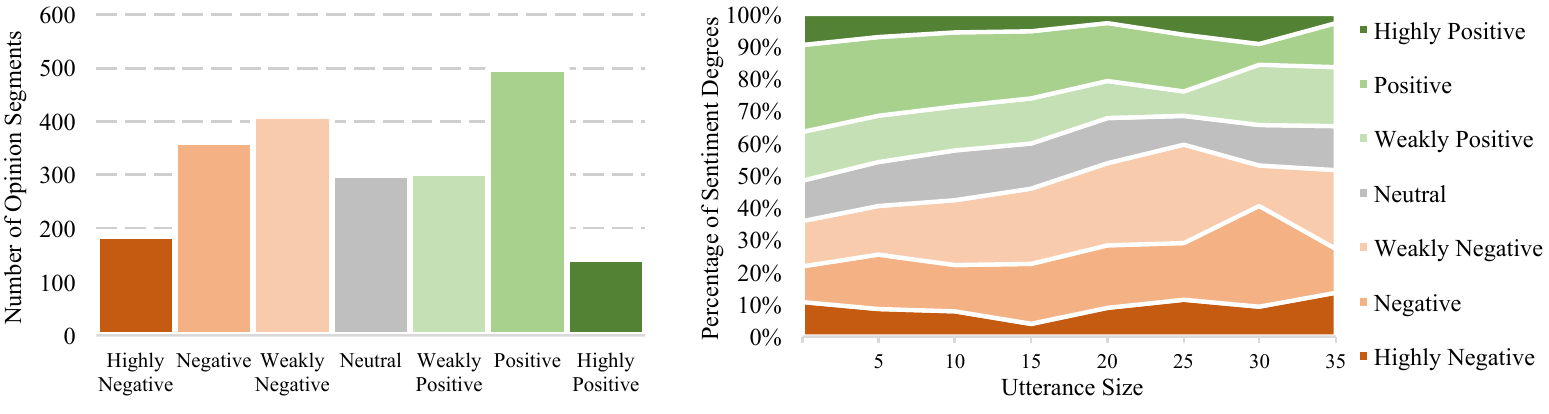}} 
    \caption{Distribution of sentiment across different opinions (left) and opinion sizes (right) in CMU-MOSI.}
    \label{fig:mosi_graph}
\end{figure*}

%% file: fig/language.tex

\begin{figure}[t!]
	\centering
{\includegraphics[width=0.9\linewidth]{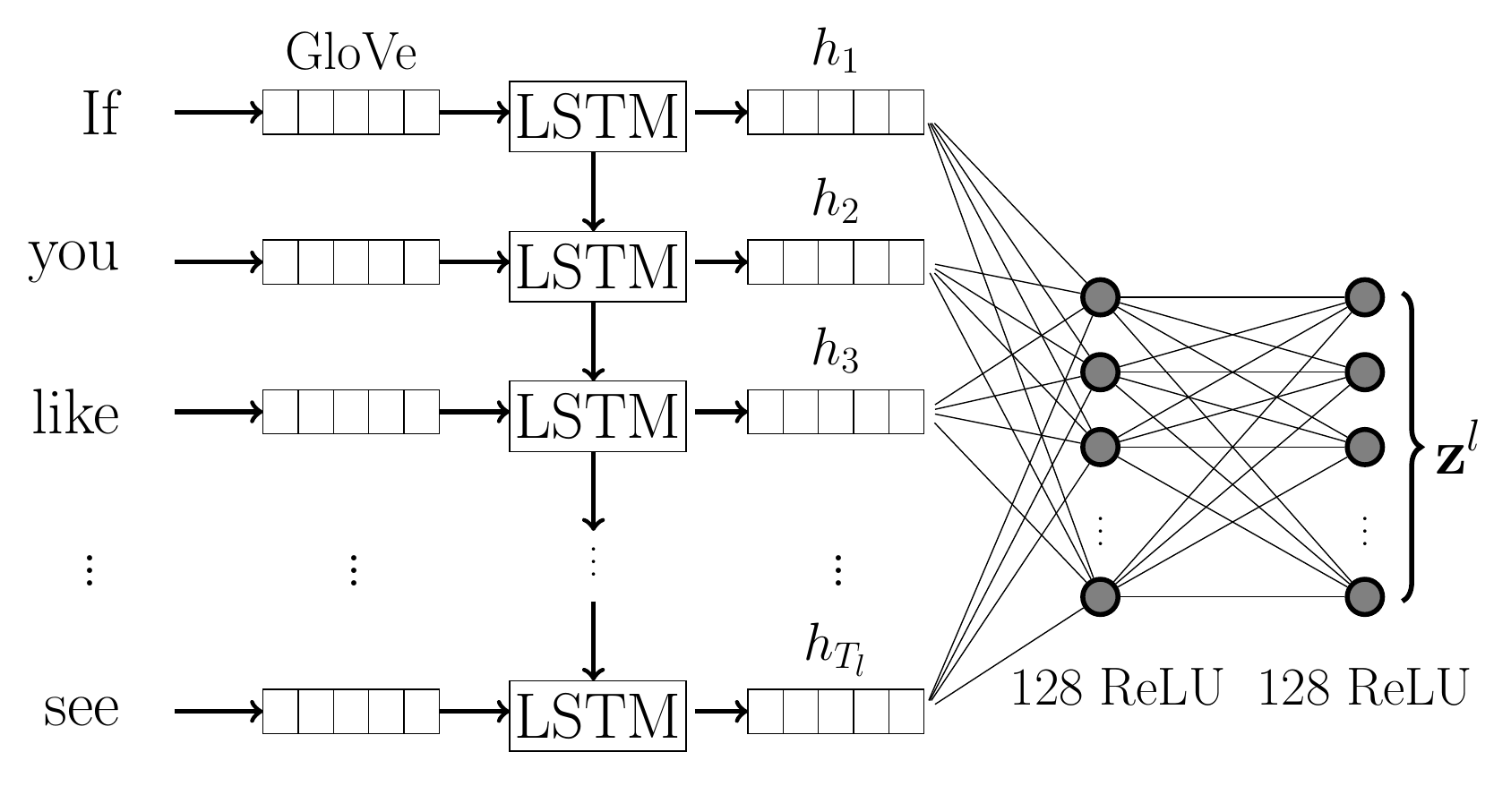}} 
    \caption{Spoken Language Embedding Subnetwork ($\mathcal{U}_l$)}
    \label{fig:lang}
\end{figure}


%% file: fig/fusion_fig.tex
\begin{figure*}[t!]
	\centering
{\includegraphics[width=\linewidth]{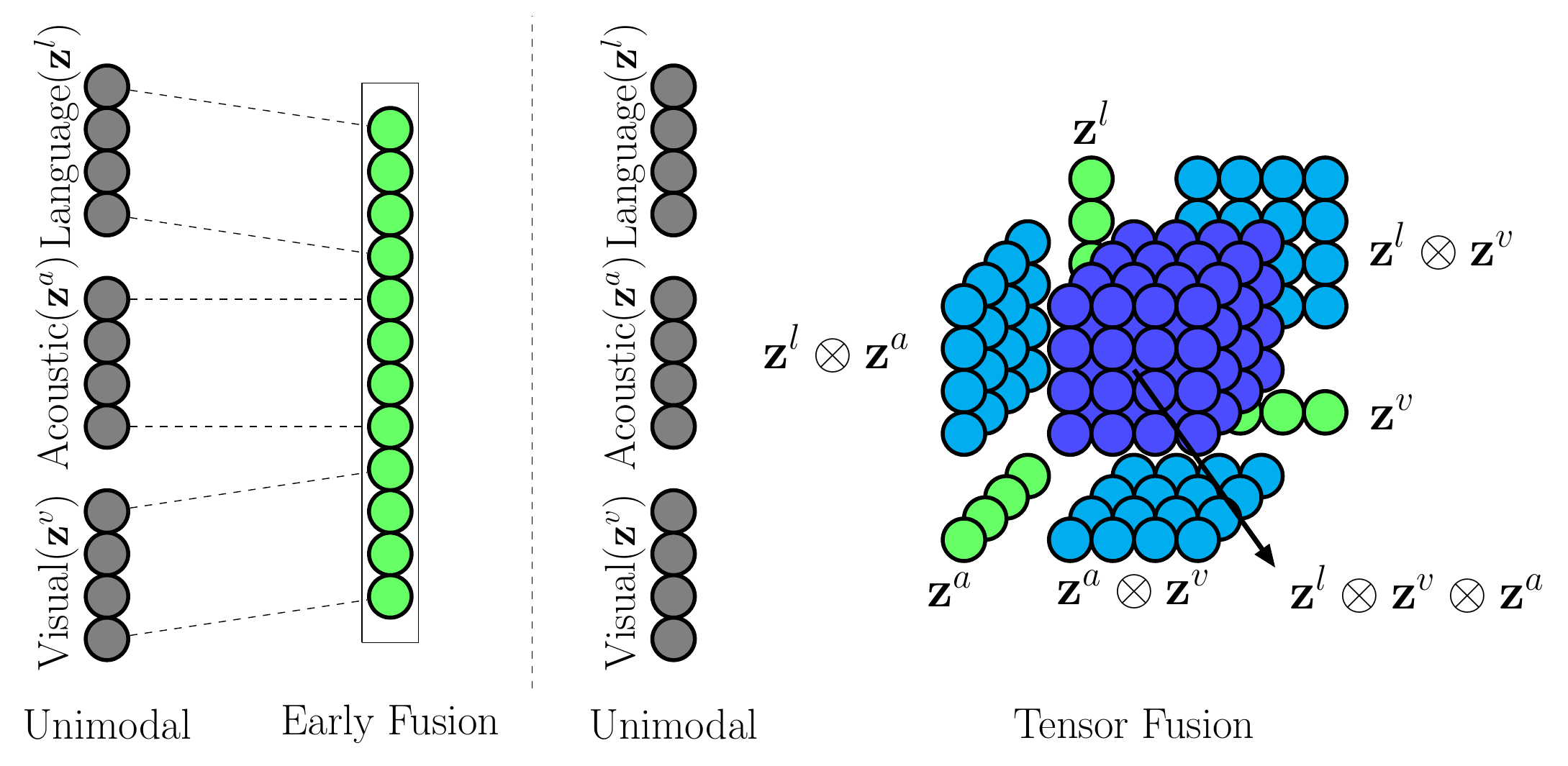}} 
    \caption{Left: Commonly used early fusion (multimodal concatenation). Right: Our proposed tensor fusion with three types of subtensors: unimodal, bimodal and trimodal.}
    \label{fig:fusion}
\end{figure*}

%% file: tables/ex1_table.tex
\newcolumntype{L}[1]{>{\raggedright\let\newline\\\arraybackslash\hspace{0pt}}m{#1}}
\newcolumntype{C}[1]{>{\centering\let\newline\\\arraybackslash\hspace{0pt}}m{#1}}
\newcolumntype{R}[1]{>{\raggedleft\let\newline\\\arraybackslash\hspace{0pt}}m{#1}}

\begin{table}
\small
\setlength\tabcolsep{0.5pt}
\begin{center}
\begin{tabular}{ L{2cm} S[table-format = 3.2, table-column-width=1.1cm] S[table-format = 3.2, table-column-width=1.1cm] S[table-format = 3.2, table-column-width=1.1cm] S[table-format = 3.4, table-column-width=1cm] S[table-format = 3.4, table-column-width=1cm] }
	\toprule
    
  	\multirow{ 2}{*}{\pbox{10cm}{Multimodal \\Baseline}} & \multicolumn{2}{c}{{Binary}} & \multicolumn{1}{c}{{5-class}} & \multicolumn{2}{c}{{Regression}} \\
    \cmidrule(lr{.5em}){2-3} \cmidrule(lr{.5em}){4-4} \cmidrule(lr{.5em}){5-6}
	&{Acc(\%)} & {F1} & {Acc(\%)} & {MAE} & {$r$} \\ 
    \midrule
  	Random & 50.2 & 48.7 & 23.9 & 1.88 & {-}\\  
    C-MKL & 73.1 & 75.2 & 35.3 & {-} & {-}\\
    SAL-CNN & 73.0 & {-} & {-} & {-} & {-}\\
  	SVM-MD & 71.6 & 72.3 & 32.0 & 1.10 & 0.53\\
    RF & 71.4 & 72.1 & 31.9 & 1.11 & 0.51\\
  	\mns & \bfseries 77.1 & \bfseries 77.9 & \bfseries 42.0 & \bfseries 0.87 & \bfseries 0.70\\
    Human & 85.7 & 87.5 & 53.9 & 0.71 & 0.82\\
    \midrule
    $\Delta^{SOTA}$ & {$\uparrow$}~~4.0 & {$\uparrow$} 2.7 & {$\uparrow$} 6.7 & {$\downarrow$}  0.23 & {$\uparrow$}  0.17\\
    \bottomrule
\end{tabular}
\end{center}
\caption{Comparison with state-of-the-art approaches for multimodal sentiment analysis. \mns \ outperforms both neural and non-neural approaches as shown by $\Delta^{SOTA}$.}\label{table:mmres}
\end{table}

%% file: tables/ex2_table2.tex
\newcolumntype{L}[1]{>{\raggedright\let\newline\\\arraybackslash\hspace{0pt}}m{#1}}
\newcolumntype{C}[1]{>{\centering\let\newline\\\arraybackslash\hspace{0pt}}m{#1}}
\newcolumntype{R}[1]{>{\raggedleft\let\newline\\\arraybackslash\hspace{0pt}}m{#1}}

\begin{table}
\small
\setlength\tabcolsep{0.5pt}
\begin{center}
\begin{tabular}{L{2cm} C{1.1cm} C{1.1cm} C{1.1cm} C{1.08cm} C{1.08cm}}
	\toprule
    \multirow{ 2}{*}{Baseline} & \multicolumn{2}{c}{Binary} & \multicolumn{1}{c}{5-class} & \multicolumn{2}{c}{Regression} \\
    \cmidrule(lr{.5em}){2-3} \cmidrule(lr{.5em}){4-4} \cmidrule(lr{.5em}){5-6}
	&Acc(\%) & F1 & Acc(\%) & MAE & $r$ \\
    \midrule
  	\mns$_{language}$ & $74.8$ & $75.6$ & $38.5$ & $0.99$ & $0.61$\\
  	\mns$_{visual}$ & $66.8$ & $70.4$ & $30.4$ & $1.13$ & $0.48$\\
  	\mns$_{acoustic}$ & $65.1$ & $67.3$ & $27.5$ & $1.23$ & $0.36$\\
    \midrule
    \mns$_{bimodal}$ & $75.2$ & $76.0$ & $39.6$ & $0.92$ & $0.65$\\
    \mns$_{trimodal}$ & $74.5$ & $75.0$ & $38.9$ & $0.93$ & $0.65$\\
    \mns$_{notrimodal}$ & $75.3$ & $76.2$ & $39.7$ & $0.919$ & $0.66$\\
    \midrule
	\mns & $\mathbf{77.1}$ & $\mathbf{77.9}$ & $\mathbf{42.0}$ & $\mathbf{0.87}$ & $\mathbf{0.70}$\\
        \mns$_{early}$ & $75.2$ & $76.2$ & $39.0$ & $0.96$ & $0.63$\\

    \bottomrule
\end{tabular}
\end{center}
\caption{Comparison of \mns \ with its subtensor variants. All the unimodal, bimodal and trimodal subtensors are important. \mns \ also outperforms early fusion. } \label{table:ex22}
\end{table}

%% file: tables/ex3_table_l.tex
\newcolumntype{L}[1]{>{\raggedright\let\newline\\\arraybackslash\hspace{0pt}}m{#1}}
\newcolumntype{C}[1]{>{\centering\let\newline\\\arraybackslash\hspace{0pt}}m{#1}}
\newcolumntype{R}[1]{>{\raggedleft\let\newline\\\arraybackslash\hspace{0pt}}m{#1}}

\begin{table}
\small
\setlength\tabcolsep{0.5pt}	
\begin{center}
\begin{tabular}{L{2cm} S[table-format = 3.2, table-column-width=1.1cm, input-symbols = {()}, group-digits = false] S[table-format = 3.2, table-column-width=1.1cm, input-symbols = {()}, group-digits = false] S[table-format = 3.2, table-column-width=1.1cm, input-symbols = {()}, group-digits = false] S[table-format = 3.4, table-column-width=1cm, input-symbols = {()}, group-digits = false] S[table-format = 3.4, table-column-width=1cm, input-symbols = {()}, group-digits = false]}
	\toprule
    \multirow{ 2}{*}{\pbox{10cm}{Language \\Baseline}} & \multicolumn{2}{c}{{Binary}} & \multicolumn{1}{c}{{5-class}} & \multicolumn{2}{c}{{Regression}} \\
    \cmidrule(lr{.5em}){2-3} \cmidrule(lr{.5em}){4-4} \cmidrule(lr{.5em}){5-6}
	&{Acc(\%)} & {F1} & {Acc(\%)} & {MAE} & {$r$} \\
    \midrule
  	\multirow{ 2}{*}{RNTN} & {-} & {-}  & {-} & {-} & {-} \\
  	& (73.7) & (73.4) & (35.2) & (0.99) & (0.59)\\ \cmidrule(lr{.5em}){2-6}
    \multirow{ 2}{*}{DAN} & 73.4 & 73.8 & \bfseries 39.2 & {-} & {-} \\
    & (68.8) &  (68.4) & (36.7) & {-} & {-} \\ \cmidrule(lr{.5em}){2-6}
    \multirow{ 2}{*}{D-CNN} & 65.5 & 66.9 & 32.0 & {-} & {-}\\
    & (62.1) & (56.4) & (32.4) & {-} & {-}\\ \cmidrule(lr{.5em}){2-6}
    CMKL-L & 71.2 & 72.4 & 34.5 & {-} & {-}\\
    SAL-CNN-L & 73.5 & {-} & {-} & {-} & {-}\\
	SVM-MD-L & 70.6 & 71.2 & 33.1 & 1.18 & 0.46\\
  	\mns $_{language}$ & \bfseries 74.8 & \bfseries 75.6 & 38.5 & \bfseries 0.98 & \bfseries 0.62\\
  	\midrule
    $\Delta_{language}^{SOTA}$ & {$\uparrow$} 1.1 & {$\uparrow$} 1.8 & {$\downarrow$} 0.7 & {$\downarrow$} 0.01 & {$\uparrow$} 0.03\\
    \bottomrule
\end{tabular}
\end{center}
\caption{Language Sentiment Analysis. Comparison of with state-of-the-art approaches for language sentiment analysis. $\Delta_{language}^{SOTA}$ shows improvement.} \label{table:lres}
\end{table}

%% file: tables/ex3_table_2.tex
\newcolumntype{L}[1]{>{\raggedright\let\newline\\\arraybackslash\hspace{0pt}}m{#1}}
\newcolumntype{C}[1]{>{\centering\let\newline\\\arraybackslash\hspace{0pt}}m{#1}}
\newcolumntype{R}[1]{>{\raggedleft\let\newline\\\arraybackslash\hspace{0pt}}m{#1}}

\begin{table}
\small
\setlength\tabcolsep{0.5pt}
\begin{center}
\begin{tabular}{L{2cm} S[table-format = 3.2, table-column-width=1.1cm] S[table-format = 3.2, table-column-width=1.1cm] S[table-format = 3.2, table-column-width=1.1cm] S[table-format = 3.4, table-column-width=1cm] S[table-format = 3.4, table-column-width=1cm]}
	\toprule
    \multirow{ 2}{*}{\pbox{10cm}{Visual \\Baseline}} & \multicolumn{2}{c}{{Binary}} & \multicolumn{1}{c}{{5-class}} & \multicolumn{2}{c}{{Regression}} \\
    \cmidrule(lr{.5em}){2-3} \cmidrule(lr{.5em}){4-4} \cmidrule(lr{.5em}){5-6}
	&{Acc(\%)} & {F1} & {Acc(\%)} & {MAE} & {$r$} \\
    \midrule
	3D-CNN & 56.1 & 58.4 & 24.9 & 1.31 & 0.26\\
    CNN-LSTM & 60.7 & 61.2 & 25.1 & 1.27 & 0.30\\
	LSTM-FA & 62.1 & 63.7 & 26.2 & 1.23 & 0.33\\
	CMKL-V & 52.6 & 58.5 & 29.3 & {-} & {-}\\
	SAL-CNN-V & 63.8 & {-} & {-} & {-} & {-}\\
  	SVM-MD-V & 59.2 & 60.1 & 25.6 & 1.24 & 0.36\\
  	\mns $_{visual}$ & \bfseries 69.4 & \bfseries 71.4 & \bfseries 31.0 & \bfseries 1.12 & \bfseries 0.50\\
    \midrule
    $\Delta_{visual}^{SOTA}$ & {$\uparrow$} 5.6 & {$\uparrow$} 7.7 & {$\uparrow$} 1.7 & {$\downarrow$} 0.11 & {$\uparrow$} 0.14\\
    \bottomrule
\end{tabular}
\end{center}
\caption{Visual Sentiment Analysis. Comparison with state-of-the-art approaches for visual sentiment analysis and emotion recognition. $\Delta_{visual}^{SOTA}$ shows the improvement.}\label{table:vres}
\end{table}

%% file: tables/ex3_table_3.tex
\newcolumntype{L}[1]{>{\raggedright\let\newline\\\arraybackslash\hspace{0pt}}m{#1}}
\newcolumntype{C}[1]{>{\centering\let\newline\\\arraybackslash\hspace{0pt}}m{#1}}
\newcolumntype{R}[1]{>{\raggedleft\let\newline\\\arraybackslash\hspace{0pt}}m{#1}}

\begin{table}
\small
\setlength\tabcolsep{0.5pt}
\begin{center}
\begin{tabular}{L{2cm} S[table-format = 3.2, table-column-width=1.1cm] S[table-format = 3.2, table-column-width=1.1cm] S[table-format = 3.2, table-column-width=1.1cm] S[table-format = 3.4, table-column-width=1cm] S[table-format = 3.4, table-column-width=1cm]}
	\toprule
    \multirow{ 2}{*}{\pbox{10cm}{Acoustic \\Baseline}} & \multicolumn{2}{c}{{Binary}} & \multicolumn{1}{c}{{5-class}} & \multicolumn{2}{c}{{Regression}} \\
    \cmidrule(lr{.5em}){2-3} \cmidrule(lr{.5em}){4-4} \cmidrule(lr{.5em}){5-6}
	&{Acc(\%)} & {F1} & {Acc(\%)} & {MAE} & {$r$} \\
    \midrule
	
  	HL-RNN & 63.4 & 64.2 & 25.9 & \bfseries 1.21 & 0.34\\
	Adieu-Net & 59.2 & 60.6 & 25.1 & 1.29 & 0.31\\
    SER-LSTM & 55.4 & 56.1 & 24.2 & 1.36 & 0.23\\
	CMKL-A & 52.6 & 58.5 & \bfseries 29.1 & {-} & {-}\\
	SAL-CNN-A & 62.1 & {-} & {-} & {-} & {-}\\
    SVM-MD-A & 56.3 & 58.0 & 24.6 & 1.29 & 0.28\\
  	\mns $_{acoustic}$ & \bfseries 65.1 & \bfseries 67.3 & 27.5 & 1.23 & \bfseries 0.36\\
    \midrule
    $\Delta_{acoustic}^{SOTA}$ & {$\uparrow$} 1.7 & {$\uparrow$} 3.1 & {$\downarrow$} 1.6 & {$\uparrow$} 0.02 & {$\uparrow$} 0.02\\
    \bottomrule
\end{tabular}
\end{center}
\caption{Acoustic Sentiment Analysis. Comparison with state-of-the-art approaches for audio sentiment analysis and emotion recognition. $\Delta_{acoustic}^{SOTA}$ shows improvement.}\label{table:vres}
\end{table}

%% file: tables/ComparisonTable.tex
\newcolumntype{C}[1]{>{\centering}m{#1}}
\begin{table*}[t!]
  \begin{center}
  \small
  \begin{tabular}{C{0.2cm} >{\small}m{5.0cm} C{1.2cm} C{1.2cm} C{1.2cm} C{1.2cm} C{1.2cm} C{1.2cm} }
    \toprule
    \# &\centering {Spoken words + \\ acoustic and visual behaviors}   &\mns-Acoustic& \mns-Visual & \mns-Language& \mns-Early& \mns  & Ground Truth\tabularnewline \midrule
    
1& \colorbox{red!50}{``You can't even tell funny jokes''} + \colorbox{red}{frowning expression} & \colorbox{black!25}{-0.375}& \colorbox{red}{-1.760}& \colorbox{red!50}{-0.558}& \colorbox{red!50}{-0.839}& \colorbox{red}{-1.661}& \colorbox{red}{-1.800} \tabularnewline \midrule
2& \colorbox{black!25}{``I gave it a B''} \  + \colorbox{green!30}{smile expression} + \colorbox{green}{excited voice} & \colorbox{green}{1.967}& \colorbox{green!30}{1.245}& \colorbox{black!25}{0.438}& \colorbox{black!25}{0.467}& \colorbox{green!30}{1.215}& \colorbox{green!30}{1.400}  \tabularnewline \midrule
3& \colorbox{green}{``But I must say those are some pretty} \colorbox{green}{big shoes to fill so I thought maybe} \colorbox{green}{it has a chance''} + \colorbox{red!50}{headshake} & \colorbox{black!25}{-0.378}& \colorbox{red!50}{-1.034}& \colorbox{green}{1.734}& \colorbox{green!30}{1.385}& \colorbox{green!30}{0.608}& \colorbox{black!25}{0.400}  \tabularnewline \midrule
4& \colorbox{black!25}{``The only actor who can really sell} \colorbox{black!25}{their lines is Erin Eckart''}+ \colorbox{red!50}{frown} + \colorbox{red!50}{low-energy voice} & \colorbox{red!50}{-0.970}& \colorbox{red!50}{-0.716}& \colorbox{black!25}{0.175}& \colorbox{black!25}{-0.031}& \colorbox{red!50}{-0.825}& \colorbox{red!50}{-1.000}  \tabularnewline \midrule

  \end{tabular}
  \captionof{table}{Examples from the CMU-MOSI dataset. The ground truth sentiment labels are between strongly negative (-3) and strongly positive (+3). For each example, we show the prediction output of the three unimodal models (\mns$_{acoustic}$, \mns$_{visual}$ and \mns$_{language}$), the early fusion model \mns$_{early}$ and our proposed \mns \ approach. \mns$_{early}$ seems to be mostly replicating language modality while our \mns \ approach successfully integrate intermodality dynamics to predict the sentiment level. 
  }  \label{table:comparison}
  \end{center}
\end{table*}